\documentclass[11pt]{article}
\usepackage{nodalida2025}
\usepackage{times}
\usepackage{url}
\usepackage{latexsym}

\usepackage[T1]{fontenc}

\usepackage[utf8]{inputenc}

\usepackage{microtype}

\usepackage{inconsolata}

\usepackage{listings}

\usepackage{hyperref}

\usepackage{tabularray}

\usepackage{algpseudocode}
\usepackage{algorithm}

\usepackage{xfrac}

\usepackage{pifont}
\newcommand{\tick}{\color{green}\ding{51}}
\newcommand{\cross}{\color{red}\ding{55}}

\newcommand{\leaderboardurl}{\url{https://scandeval.com}}

\aclfinalcopy % Uncomment this line for the final submission

\title{
    Encoder vs Decoder: Comparative Analysis of Encoder and Decoder Language Models on
    Multilingual NLU Tasks
}

\author{
    Dan Saattrup Nielsen \\
    The Alexandra Institute \\
    \small\texttt{dan.nielsen@alexandra.dk} \\\And
    Kenneth Enevoldsen \\
    University of Aarhus \\
    \small\texttt{kenneth.enevoldsen@cas.au.dk} \\\And
    Peter Schneider-Kamp \\
    University of Southern Denmark \\
    \small\texttt{petersk@imada.sdu.dk}
}

\date{}

\begin{document}
\maketitle
\begin{abstract}
    This paper explores the performance of encoder and decoder language models on
    multilingual Natural Language Understanding (NLU) tasks, with a broad focus on
    Germanic languages. Building upon the ScandEval benchmark, initially restricted to
    evaluating encoder models, we extend the evaluation framework to include decoder
    models. We introduce a method for evaluating decoder models on NLU tasks and apply
    it to the languages Danish, Swedish, Norwegian, Icelandic, Faroese, German, Dutch,
    and English. Through a series of experiments and analyses, we also address research
    questions regarding the comparative performance of encoder and decoder models, the
    impact of NLU task types, and the variation across language resources. Our findings
    reveal that encoder models can achieve significantly better NLU performance than
    decoder models despite having orders of magnitude fewer parameters. Additionally,
    we investigate the correlation between decoders and task performance via a UMAP
    analysis, shedding light on the unique capabilities of decoder and encoder models.
    This study contributes to a deeper understanding of language model paradigms in NLU
    tasks and provides valuable insights for model selection and evaluation in
    multilingual settings.
\end{abstract}

\section{Introduction}

Language models have attained remarkable Natural Language Understanding (NLU)
performance, both with encoder-based architectures like BERT \cite{devlin2018bert} and
and decoder-based architectures like GPT-3 \cite{brown2020language}. The encoder models
have excelled in capturing contextual information for downstream tasks through masked
language modeling objectives, while decoder models have shown strong generative
capabilities by autoregressively predicting subsequent tokens based on preceding
context.

Since the ``ChatGPT boom'' in 2023, the research community has been increasingly
focused on decoder models \cite{zhao2023survey} for both Natural Language Generation
(NLG) and NLU tasks. However, few studies have systematically compared the performance
of encoder and decoder models across a diverse range of NLU tasks, and the studies that
exist have primarily focused on English. This leaves a gap in our understanding of how
the two language model paradigms perform in multilingual settings across different
languages and tasks.

\citet{nielsen2023scandeval} introduced the ScandEval benchmark and evaluated encoder
language models on four different natural language understanding tasks in Danish,
Swedish, Norwegian (Bokmål and Nynorsk), Icelandic and Faroese. In this paper, we
bridge this gap by extending the ScandEval benchmark to encompass the evaluation of
decoder models on multilingual NLU tasks, as well as expanding the language resources
to include German, Dutch and English.

Our \textbf{main research question} is

\begin{quote}
    \textit{Which language model paradigm is better suited for NLU?}
\end{quote}

We will answer this question with the languages Danish, Swedish, Norwegian, Icelandic,
Faroese, German, Dutch and English as a case study. To concretise our main question, we
will study the following research questions in this paper:

\begin{enumerate}
    \item[(Q1)] Can state-of-the-art finetuned encoder models achieve significantly
      better NLU performance than state-of-the-art decoder models?
    \item[(Q2)] Does the answer to (Q1) depend on the type of NLU task?
    \item[(Q3)] Does the answer to (Q1) vary along the language resource spectrum, from
      low- to high-resource?
\end{enumerate}

Our main contributions of this paper are the following:

\begin{enumerate}
    \item We extend the ScandEval benchmarking framework with few-shot evaluation of
        decoder models and release this extension open-source.
    \item We extend the languages supported by the ScandEval benchmarking framework by
        German, Dutch and English. Together with Danish, Swedish, Norwegian, Icelandic
        and Faroese, ScandEval now provides coverage of all Germanic languages except
        Afrikaans and the Frisian languages.
    \item We evaluate an extensive suite of both encoder and decoder models on NLU
        tasks in all of the supported languages and publish these on public
        leaderboards.
    \item We give a positive answer to (Q1), showing that encoder models achieve
      significantly better NLU performance than encoder models in several languages.
      This depends on the language in question however, giving a partially positive
      answer to (Q3).
    \item We also show that the decoder models are heavily biased towards the question
        answering task (even models that are not instruction tuned), and a UMAP
        analysis shows that the performance distribution of decoder models follow a
        different ``path'' than encoder models, from the worst to best performing
        models. This gives a positive answer to (Q2).
\end{enumerate}

\section{Related Work}

\subsection{Comparing Encoder and Decoder Models}

There has been a number of studies in recent years comparing encoder models to decoder
models. \citet{zhong2023can} compared GPT-3.5-turbo (January 2023 version) to
(finetuned versions of) the base and large versions of BERT \cite{devlin2018bert} and
RoBERTa \cite{liu2019roberta} on the English GLUE benchmark \cite{wang-etal-2018-glue}.
They find that GPT-3.5-turbo is on average on par with the base-sized encoder models,
but falls short of the large-sized ones. They also note that despite being on par with
the base-sized models, there is a big discrepancy between the models on individual
tasks, with GPT-3.5-turbo for instance being better on the inference tasks while being
worse on the paraphrase tasks. We note however that they only evaluate the decoder
model in a zero-shot setting, and furthermore they only evaluate the models on 25
samples for each class in the development split, leading to a potential lack of
robustness in their evaluation.

\citet{wang2023chatgpt} compares GPT-3.5-turbo (January 2023 version) to a finetuned
version of the base-sized BERT model on 18 English benchmark datasets related to
sentiment analysis. Like \citet{zhong2023can}, they find that the zero-shot
performance of GPT-3.5-turbo is on par with the base-sized BERT model, and that the
few-shot performance of GPT-3.5-turbo (with 27 few-shot examples) is slightly better
than BERT, on average. Their test sets contained, on average, 538 samples, which is a
significant improvement over \citet{zhong2023can}. However, the narrow focus on the
evaluation tasks as well as only benchmarking a single encoder and decoder model makes
it hard to generalise the results to other tasks and models.

\citet{kocon2023chatgpt} built a benchmark suite of 25 tasks, where 21 of these tasks
are classification tasks (binary, multi-class and multi-label), 3 being question
answering tasks and the last one being a token classification task. Two of the
classification tasks are in Polish and the rest in English. They compare the
zero-shot and few-shot performance of GPT-3.5-turbo (January 2023 version) to the
state-of-the-art encoder performance on each task. GPT-3.5-turbo is generally found to
be worse than state-of-the-art encoder models. They also evaluate GPT-4 on five of the
tasks (inference, question-answering and emotion datasets), and only find GPT-4 to be
marginally better than GPT-3.5-turbo, still far off the encoder models.

\citet{qiu2024chatgpt} compare GPT-3.5-turbo (January 2023 version) to a finetuned
version of the base-sized BERT model on three manually curated English multi-class
classification datasets with 19, 12 and 7 test samples, respectively, where they find
that the BERT model performs marginally better than GPT-3.5-turbo in a few-shot
setting (and that the zero-shot performance is significantly worse). The tiny test
sets make it hard to generalise the results, however.

\subsection{Benchmarks of Generative Language Models}

In recent times, several benchmarks of generative language models have been introduced.
The major ones are EleutherAI's Evaluation Harness \cite{eval-harness}, Hugging Face's
Open LLM Leaderboard \cite{huggingfaceOpenLeaderboard} which uses the Evaluation
Harness as evaluation engine, and Stanford University's HELM
\cite{bommasani2023holistic}. These are firstly all English-only benchmarks, making it
hard to generalise the results to other languages, and they only include point
estimates of the dataset performance, and thus do not necessarily provide a robust
assessment of the models. Further, these benchmarks are exclusively for decoder models,
and thus does not provide a way to compare encoders with decoders.

There has been several language-specific benchmarks introduced as well. NorBench
\cite{samuel2023norbench} is a collection of Norwegian evaluation datasets moreso than a
dedicated evaluation framework. Further, several datasets in this collection (NorQuAD,
NoReC and NorNE) are already part of ScandEval. SuperLim
\cite{berdivcevskis2023superlim} falls into the same category for Swedish. DUMB
\cite{de2023dumb} is a Dutch benchmarking framework, which is only focused on encoder
models. Danoliterate \cite{holm2024danoliterate} is a Danish benchmarking framework
which is solely focused on evaluating decoder models, and whose datasets largely overlap
with the Danish datasets in ScandEval, albeit with a different evaluation methodology.
Aside from language modelling performance, the Danoliterate benchmark also measures
calibration, efficiency, toxicity and fairness. While the development of
language-specific benchmarks is important, it leads to too little overview of trends
across benchmarks and languages and incentivises model development focused on
monolingual models ignoring a potential broader appeal. ScandEval provides a unified and
robust approach for comparison across model categories and Germanic languages.

Benchmarking is not the only way to evaluate language models. A new ``arena approach''
has been popularised by the LMSYS Arena \cite{chiangchatbot}, where users can submit a
prompt and get two responses from two anonymised models at random, and have to evaluate
the responses. The Arena is predominantly used for English, but also currently supports
six other languages. This approach is a promising way to evaluate language models, but
we fear that it is not as suitable for low-resource languages due to the need of many
volunteers to evaluate the responses.

Lastly, the Scandinavian Embedding Benchmark \cite{enevoldsen2024scandinavian}
complements ScandEval and focuses on evaluating embedding models on a wide range of
tasks in the Scandinavian languages.

\section{Datasets}
In this section we present the datasets that we are evaluating the models on, all of
which are now included in the ScandEval framework. We should note that these datasets
either (a) already existed prior to this publication or (b) are small extensions of
existing datasets. An overview of all the datasets can be found in
Table~\ref{tab:datasets}.

\begin{table*}
    \centering
    \small
    \begin{tabular}{lllccc}
        \hline
        \textbf{Dataset} & \textbf{Language} &
        \textbf{\#Train} & \textbf{\#Val} & \textbf{\#Test} & \textbf{\#Shots} \\
        \hline
        \textbf{NER} \\
            \hspace{3mm} DANSK \cite{enevoldsen2024dansk} & Danish  & 1,024 & 256 & 1,024 & 8 \\
            \hspace{3mm} SUC 3.0 \cite{gustafson2006manual} & Swedish  & 1,024 & 256 & 2,048 & 8 \\
            \hspace{3mm} NorNE-nb \cite{jorgensen-etal-2020-norne} & Norwegian Bokmål  & 1,024 & 256 & 2,048 & 8 \\
            \hspace{3mm} NorNE-nn \cite{jorgensen-etal-2020-norne} & Norwegian Nynorsk  & 1,024 & 256 & 2,048 & 8 \\
            \hspace{3mm} MIM-GOLD-NER \cite{ingolfsdottir2020named} & Icelandic  & 1,024 & 256 & 2,048 & 8 \\
            \hspace{3mm} FoNE \cite{snaebjarnarson2023transfer} & Faroese  & 1,024 & 256 & 2,048 & 8 \\
            \hspace{3mm} GermEval \cite{benikova2014nosta} & German  & 1,024 & 256 & 1,024 & 8 \\
            \hspace{3mm} CoNLL-nl \cite{sang2002introduction} & Dutch  & 1,024 & 256 & 1,024 & 8 \\
            \hspace{3mm} CoNLL-en \cite{sang2003introduction} & English  & 1,024 & 256 & 2,048 & 8 \\
        \textbf{Sentiment Classification} \\
            \hspace{3mm} Angry Tweets \cite{pauli2021danlp} & Danish  & 1,024 & 256 & 2,048 & 12 \\
            \hspace{3mm} SweReC \cite{svensson2017sentiment} & Swedish  & 1,024 & 256 & 2,048 & 12 \\
            \hspace{3mm} NoReC \cite{velldal-etal-2018-norec} & Norwegian  & 1,024 & 256 & 2,048 & 12 \\
            \hspace{3mm} SB10k \cite{cieliebak2017twitter} & German  & 1,024 & 256 & 1,024 & 12 \\
            \hspace{3mm} Dutch Social \cite{gupta2022dutchsocial} & Dutch  & 1,024 & 256 & 1,024 & 12 \\
            \hspace{3mm} SST5 \cite{socher2013recursive} & English  & 1,024 & 256 & 2,048 & 12 \\
        \textbf{Linguistic Acceptability} \\
            \hspace{3mm} ScaLA-da \cite{nielsen2023scandeval} & Danish & 1,024 & 256 & 2,048 & 12 \\
            \hspace{3mm} ScaLA-sv \cite{nielsen2023scandeval} & Swedish & 1,024 & 256 & 2,048 & 12 \\
            \hspace{3mm} ScaLA-nb \cite{nielsen2023scandeval} & Norwegian Bokmål & 1,024 & 256 & 2,048 & 12 \\
            \hspace{3mm} ScaLA-nn \cite{nielsen2023scandeval} & Norwegian Nynorsk & 1,024 & 256 & 2,048 & 12 \\
            \hspace{3mm} ScaLA-is \cite{nielsen2023scandeval} & Icelandic & 1,024 & 256 & 2,048 & 12 \\
            \hspace{3mm} ScaLA-fo \cite{nielsen2023scandeval} & Faroese & 1,024 & 256 & 1,024 & 12 \\
            \hspace{3mm} ScaLA-de \cite{nielsen2023scandeval} & German & 1,024 & 256 & 2,048 & 12 \\
            \hspace{3mm} ScaLA-nl \cite{nielsen2023scandeval} & Dutch & 1,024 & 256 & 2,048 & 12 \\
            \hspace{3mm} ScaLA-en \cite{nielsen2023scandeval} & English & 1,024 & 256 & 2,048 & 12 \\
        \textbf{Question Answering} \\
            \hspace{3mm} ScandiQA-da \cite{nielsen2023scandeval} & Danish & 1,024 & 256 & 2,048 & 4 \\
            \hspace{3mm} ScandiQA-sv \cite{nielsen2023scandeval} & Swedish & 1,024 & 256 & 2,048 & 4 \\
            \hspace{3mm} NorQuAD \cite{ivanova2023norquad} & Norwegian Bokmål & 1,024 & 256 & 2,048 & 2 \\
            \hspace{3mm} NQiI \cite{snaebjarnarson2022natural} & Icelandic & 1,024 & 256 & 1,024 & 4 \\
            \hspace{3mm} GermanQuAD \cite{moller2021germanquad} & German & 1,024 & 256 & 2,048 & 4 \\
            \hspace{3mm} SQuAD-nl \cite{havinga2023squadv2dutch} & Dutch & 1,024 & 256 & 2,048 & 4 \\
            \hspace{3mm} SQuAD \cite{rajpurkar2016squad} & English & 1,024 & 256 & 2,048 & 4 \\
        \hline
    \end{tabular}
    \caption{
        All the datasets used in the NLU evaluation. Note that these have been re-sized
        and do not represent the sizes of the original dataset.
    }
    \label{tab:datasets}
\end{table*}

\subsection{Named Entity Recognition}
For Norwegian, Swedish and Icelandic we use the NorNE \cite{jorgensen-etal-2020-norne},
SUC 3.0 \cite{gustafson2006manual}, MIM-GOLD-NER \cite{ingolfsdottir2020named}
datasets, which were already included in the ScandEval framework. For Faroese we
replace the previous WikiANN-fo dataset \cite{rahimi2019massively} with the new human
annotated FoNE dataset \cite{snaebjarnarson2023transfer}. We also replace the previous
DaNE dataset \cite{hvingelby2020dane} with the new DANSK dataset
\cite{enevoldsen2024dansk} covering a wider variety of domains. For German, Dutch and
English we add the established NER datasets GermEval \cite{benikova2014nosta}, the
Dutch part of CoNLL-2002 \cite{sang2002introduction}, and the English CoNLL-2003
\cite{sang2003introduction}.

\subsection{Sentiment Classification}
We re-use the sentiment classification datasets AngryTweets \cite{pauli2021danlp},
NoReC \cite{velldal-etal-2018-norec} and SweReC \cite{svensson2017sentiment}, for
Danish, Norwegian and Swedish, respectively. For German, Dutch and English we add the
existing datasets SB10k \cite{cieliebak2017twitter}, Dutch Social
\cite{gupta2022dutchsocial} and SST5 \cite{socher2013recursive}. We convert SST5 to the
standardised trinary (negative, neutral, positive) format by converting the ``very
negative'' and ``very positive'' labels to ``negative'' and ``positive'', respectively.

\subsection{Linguistic Acceptability}
For linguistic acceptability we re-use the ScaLA datasets for all the Scandinavian
languages, and extend the ScaLA datasets by applying the ScaLA method from
\citet{nielsen2023scandeval} to German, Dutch and English by using the German
\cite{mcdonald2013universal}, Dutch \cite{van2002alpino} and English
\cite{zeldes2017gum} dependency treebanks.

\subsection{Extractive Question Answering}
Here we use the ScandiQA dataset \cite{nielsen2023scandeval} for Danish and Swedish,
but replace the manually translated Norwegian ScandiQA dataset with the new curated
NorQuAD dataset \cite{ivanova2023norquad}. We further add the new Natural Questions in
Icelandic dataset \cite{snaebjarnarson2022natural} for Icelandic. For German and
English we add the existing extractive question-answering datasets GermanQuAD
\cite{moller2021germanquad} and SQuAD \cite{rajpurkar2016squad}, respectively. For
Dutch we add the machine translated version of SQuAD to Dutch
\cite{havinga2023squadv2dutch}.

\section{Methodology}

\subsection{Formulating NLU Tasks as Generative Tasks}
In this section we describe how we rephrase the NLU tasks as text-to-text tasks, which makes it possible to evaluate generative models on the tasks. We formulate all the tasks as few-shot tasks, generally formatted as follows:

\begin{quote}
    \footnotesize
    [prefix prompt]\\

    [document prefix]: [document]

    [label prefix]: [label]\\

    (...)\\

    [document prefix]: [document]

    [label prefix]:
\end{quote}

We found that the separation of the few-shot examples with double newlines makes it
easier to know when to stop the generation - for the same reason, we ensure that there
are no double newlines in any of the documents. See the prompts used for the English
datasets in Table~\ref{tab:english-prompts}; a full table of the prompts used for all
the tasks in all the languages can be found in Appendix \ref{apx:datasets}.

For the sentiment classification task, we simply have the models generate translations
of the three labels (positive, negative and neutral). For the linguistic acceptability
task, also a text classification task, we use the translations of ``yes'' and ``no'' as
the two labels, corresponding to whether the document is grammatically correct or not.
For the extractive question answering task, we have the model output the answer
directly. For this task we found that changing the label prefix from ``Answer'' to
``Answer in max 3 words'' resulted in a drastic improvement, due to many of the answers
of instruction tuned models starting with unnecessary text akin to ``The answer is''.
Lastly, for the named entity recognition task, we require the output to be a JSON
dictionary \cite{json}, with keys being the translated named entity tags, and values
being lists of named entities of that category. To ensure that we are not biasing the
evaluation toward models knowing the JSON format, we employ structured generation using
the \texttt{outlines} package \cite{outlines}, which modifies the logits outputted by
the model to ensure that the output is always a valid JSON dictionary in the
aforementioned format.

\begin{table*}[]
    \centering
    \small
    \begin{tblr}{Q[3cm,valign=m]|Q[7cm,valign=m]X[j,valign=m]}
        \hline
        \textbf{Task} & \textbf{Prefix Prompt} & \textbf{Example Prompt} \\
        \hline
        Named entity recognition & Below are sentences and JSON dictionaries with the named entities that occur in the given sentence. & Sentence: [text]\newline Named entities: [label] \\
        Sentiment classification & The following are tweets are their sentiment, which can be 'positive', 'neutral' or 'negative'. & Tweet: [text]\newline Sentiment: [label] \\
        Linguistic acceptability & The following are sentences and whether they are grammatically correct. & Sentence: [text]\newline Grammatically correct: [label] \\
        Question answering & The following are texts with accompanying questions and answers. & Text: [text]\newline Question: [question]\newline Answer in max 3 words: [label] \\
        \hline
    \end{tblr}
    \caption{The English prompt templates used for the datasets. See all the prompt
    templates in Appendix \ref{apx:datasets}.}
    \label{tab:english-prompts}
\end{table*}

\subsection{Evaluation Methodology}
We keep the evaluation methodology for the generative models to be as close to the
methodology for encoder models in \citet{nielsen2023scandeval}. We think of the
few-shot examples as analogous to training examples for encoder models. Indeed, as
\citet{von2023uncovering} shows, this assumption is theoretically grounded. We thus
evaluate the models 10 times, where on each iteration we sample few-shot examples at
random from the training split, and we evaluate the model on a bootstrapped version of
the test split. As with the encoder models, this allows us to take into account more
noise in evaluation process, resulting in more robust evaluation scores.

The number of few-shot examples for each dataset was determined on a heuristic basis,
where we wanted to include as many examples as possible, while making sure that the
token count was sufficiently low to not bias the evaluation towards models with a longer
context length. All the NER, sentiment classification and linguistic acceptability
datasets have prompt sizes around 1,000 tokens with the Mistral-7B-v0.1 tokeniser
\cite{jiang2023mistral}, with the question answering datasets having around 2,000
tokens. This is also the reason for the discrepancy with the NorQuAD dataset, as the
samples are much longer than the other question answering datasets.

\subsection{Score Aggregation Method}
From the raw scores of the 10 evaluations per dataset, we need to aggregate
the model scores into a single score. We want an aggregation method that satisfies the
following criteria:

\begin{enumerate}
    \item \textbf{Task Fairness:} Each task should be weighted equally.
    \item \textbf{Comparison:} If we evaluate models in multiple languages, then
        it should be possible to meaningfully compare the language scores of these
        models with each other.
    \item \textbf{Robustness:} If two models do not have a significantly different
        score on a dataset, then the aggregated score should reflect this.
    \item \textbf{Magnitude Preservation:} The magnitude of the difference between the
        dataset score of two models should be reflected in the aggregated score.
    \item \textbf{Minimal Change:} Adding a new model should minimally affect the
        aggregated scores of the other models.
\end{enumerate}

Before we introduce our chosen aggregation method, we will briefly discuss some common
aggregation methods and how they do not satisfy the criteria.

The \textbf{mean score} is the most common aggregation method, which would simply be
the mean of the 10 scores for each dataset, and then the mean of the dataset scores for
each task. This method does not satisfy the Task Fairness criterion, as it does not
take into account that metrics have different ranges and variances. The Comparison
criterion is also not satisfied, as datasets vary from language to language, with some
datasets being more difficult than others. It \textit{does}, however, satisfy the
Robustness, Magnitude Preservation and Minimal Change criteria.

The \textbf{mean rank} is another common aggregation method, where we compute the rank
of each model on each dataset, and then take the mean of the ranks. This method
satisfies the Task Fairness criterion, as it re-casts the scores into a common
comparable framework, which therefore weights each task equally. For the same reason,
it also satisfies the Comparison criterion (it is important here that we evaluate all
the models on all the languages for this to be satisfied). It does not satisfy the
Robustness and Magnitude Preservation criteria, by definition of rank. It partially
satisfies the Minimal Change criterion, since it only affects the scores of the models
which are worse than the new model.

We thus see that the mean score and mean rank methods satisfy a disjoint set of the
criteria, but that they together satisfy all the criteria. Based on this observation,
we introduce the \textbf{mean rank score} method, defined as follows. For each dataset,
we start by sorting the models by their mean score on the dataset. As with a rank, we
assign the best model with rank score 1. For the next best model, we conduct a
one-tailed Welch's t-test to see if the next best model is significantly worse than the
first model ($p < 0.05$). If so, we compute the absolute difference between the mean
score of the two models, and divide that by the standard deviation of all the mean
scores of the models on the dataset.

We then add this to the rank score of the first model. We continue this process for all
the models to get the rank scores for the dataset, and to compute the overall score for
the model, we take the mean of the rank scores for the datasets. An overview of this
aggregation method can be found in Appendix \ref{apx:mean_rank_score}. We note that the
mean rank score has an intuitive interpretation: it is the average number of standard
deviations from the best scoring model (+1).

This metric satisfies Task Fairness since we normalise all the scores by dividing by
the standard deviation of the dataset scores. The Robustness criterion is satisfied due
to our use of a one-tailed Welch's t-test. The Magnitude Preservation criterion is also
satisfied, as the magnitude of the difference between the dataset score of two models
is reflected in the rank score. It also satisfies Comparison, as we compare the models
on a common scale (same argument as the mean rank method). Finally, the Minimal Change
criterion is partially satisfied, as adding new models only minimally changes the score
of existing models. Concretely, adding new scores will affect the standard deviation
normalising factor (this effect tends to zero as the number of models grows, however),
and if the model beats all the other models then all the scores will be affected, due
to the relative nature of the metric.

\section{Analysis}

\subsection{Comparative Performance Analysis on High- and Low-resource Languages}

Excerpts of the English, Danish and Icelandic leaderboards can be found in
Table~\ref{tab:english-leaderboard}, Table~\ref{tab:danish-leaderboard} and
Table~\ref{tab:icelandic-leaderboard}, respectively. We found that these three represent
three main categories of languages with respect to the open-closed source divide.
Similar excerpts for the remaining languages (Swedish, Norwegian, Faroese, German and
Dutch) can be found in Appendix \ref{apx:leaderboard_excerpts}. The full leaderboards
for all the languages can be found at \leaderboardurl.

From the English results we see that the state-of-the-art decoder model GPT-4-0613
\cite{achiam2023gpt} is still outperformed by the DeBERTa-v3-large and DeBERTa-v3-base
models \cite{he2020deberta} as well as the ELECTRA-base model \cite{clark2020electra}.
Here GPT-4-0613 is, on average, 0.44 standard deviations worse than the best model. The
same pattern is seen for Norwegian, Dutch, German and Faroese; see Appendix
\ref{apx:leaderboard_excerpts} for the corresponding leaderboard excerpts.

In contrast, on the Danish leaderboard, the top-3 models are all decoder models, with
GPT-4-0613 and GPT-4-1106-preview \cite{openai2023gpt4turbo} in the lead, followed by
the closed-source DanskGPT-Chat-Llama3-70B model from
Syv.AI\footnote{\url{https://www.syv.ai/}}, being a continuation of the Llama-3-70B
model \cite{llama3modelcard}. The GPT-4-0613 model is, on average, 0.24 standard
deviations from the best model. Similar results were found with Swedish; see Appendix
\ref{apx:leaderboard_excerpts} for the corresponding leaderboard excerpt.

Lastly, for Icelandic, we see that the encoders and decoders are tied in performance,
with the mDeBERTa-v3-base model and the GPT-4-1106-preview model
being the top models. The GPT-4-1106-preview model is, on average, 0.24 standard
deviations from the best model. We note that Icelandic is the \textit{only} language
where the switch from GPT-4 (gpt-4-0613) to GPT-4-turbo (gpt-4-1106-preview) resulted
in a significant \textit{increase} in performance. We speculate that this is due to
the collaboration between OpenAI and Iceland \cite{openai2023iceland}.

We can thus give an affirmative answer to research question (Q1), showing that encoder
models \textit{can} achieve significantly better NLU performance than decoder models,
even though they have an order of magnitude fewer model parameters. For (Q3), we see
that this varies between languages, but without being correlated to the language
resource spectrum.

\begin{table}[]
    \centering
    \scriptsize
    \begin{tabular}{l|cc}
        \hline
        \textbf{Model ID} & \textbf{Decoder} & \textbf{Score ($\downarrow$)} \\
        \hline
        microsoft/deberta-v3-large & \cross & 1.09 \\
        microsoft/deberta-v3-base & \cross & 1.29 \\
        google/electra-base-discriminator & \cross & 1.39 \\
        gpt-4-0613 & \tick & 1.44 \\
        FacebookAI/roberta-large & \cross & 1.46 \\
        FacebookAI/roberta-base & \cross & 1.51 \\
        microsoft/mdeberta-v3-base & \cross & 1.53 \\
        gpt-4-1106-preview & \tick & 1.54 \\
        gpt-4o-2024-05-13 & \tick & 1.64 \\
        AI-Sweden-Models/roberta-large-1160k & \cross & 1.64 \\
        gpt-3.5-turbo-0613 & \tick & 1.78 \\
        mistralai/Mistral-7B-v0.1 & \tick & 1.91 \\
        \hline
    \end{tabular}
    \caption{Excerpt of the English ScandEval leaderboard.}
    \label{tab:english-leaderboard}
\end{table}

\begin{table}[]
    \centering
    \scriptsize
    \begin{tabular}{l|cc}
        \hline
        \textbf{Model ID} & \textbf{Decoder} & \textbf{Score ($\downarrow$)} \\
        \hline
        gpt-4-0613 & \tick & 1.24 \\
        gpt-4-1106-preview & \tick & 1.25 \\
        syvai/danskgpt-chat-llama3-70b & \tick & 1.29 \\
        AI-Sweden-Models/roberta-large-1160k & \cross & 1.39 \\
        danish-foundation-models/encoder-large-v1 & \cross & 1.40 \\
        meta-llama/Meta-Llama-3-70B & \tick & 1.40 \\
        AI-Sweden-Models/Llama-3-8B-instruct & \tick & 1.44 \\
        gpt-4o-2024-05-13 & \tick & 1.46 \\
        ltg/norbert3-large & \cross & 1.50 \\
        NbAiLab/nb-bert-large & \cross & 1.54 \\
        vesteinn/DanskBERT & \cross & 1.56 \\
        google/rembert & \cross & 1.61 \\
        intfloat/multilingual-e5-large & \cross & 1.62 \\
        gpt-3.5-turbo-0613 & \tick & 1.68 \\
        FacebookAI/xlm-roberta-large & \cross & 1.71 \\
        \hline
    \end{tabular}
    \caption{Excerpt of the Danish ScandEval leaderboard.}
    \label{tab:danish-leaderboard}
\end{table}

\begin{table}[]
    \centering
    \scriptsize
    \begin{tabular}{l|cc}
        \hline
        \textbf{Model ID} & \textbf{Decoder} & \textbf{Score ($\downarrow$)} \\
        \hline
        microsoft/mdeberta-v3-base & \cross & 1.33 \\
        gpt-4-1106-preview & \tick & 1.34 \\
        gpt-4o-2024-05-13 & \tick & 1.43 \\
        vesteinn/ScandiBERT-no-faroese & \cross & 1.48 \\
        google/rembert & \cross & 1.57 \\
        vesteinn/XLMR-ENIS & \cross & 1.59 \\
        gpt-4-0613 & \tick & 1.79 \\
        mideind/IceBERT-large & \cross & 1.85 \\
        vesteinn/FoBERT & \cross & 1.87 \\
        meta-llama/Meta-Llama-3-70B & \tick & 2.03 \\
        FacebookAI/xlm-roberta-large & \cross & 2.34 \\
        gpt-3.5-turbo-0613 & \tick & 2.51 \\
        mistralai/Mistral-7B-v0.1 & \tick & 2.96 \\
        \hline
    \end{tabular}
    \caption{Excerpt of the Icelandic ScandEval leaderboard.}
    \label{tab:icelandic-leaderboard}
\end{table}

\subsection{Task Analysis}

In this section we analyse our research question (Q2), asking whether the NLU
performance results from the previous section is dependent on the type of NLU task.

\begin{figure*}[h]
    \centering
    \includegraphics[width=1\textwidth]{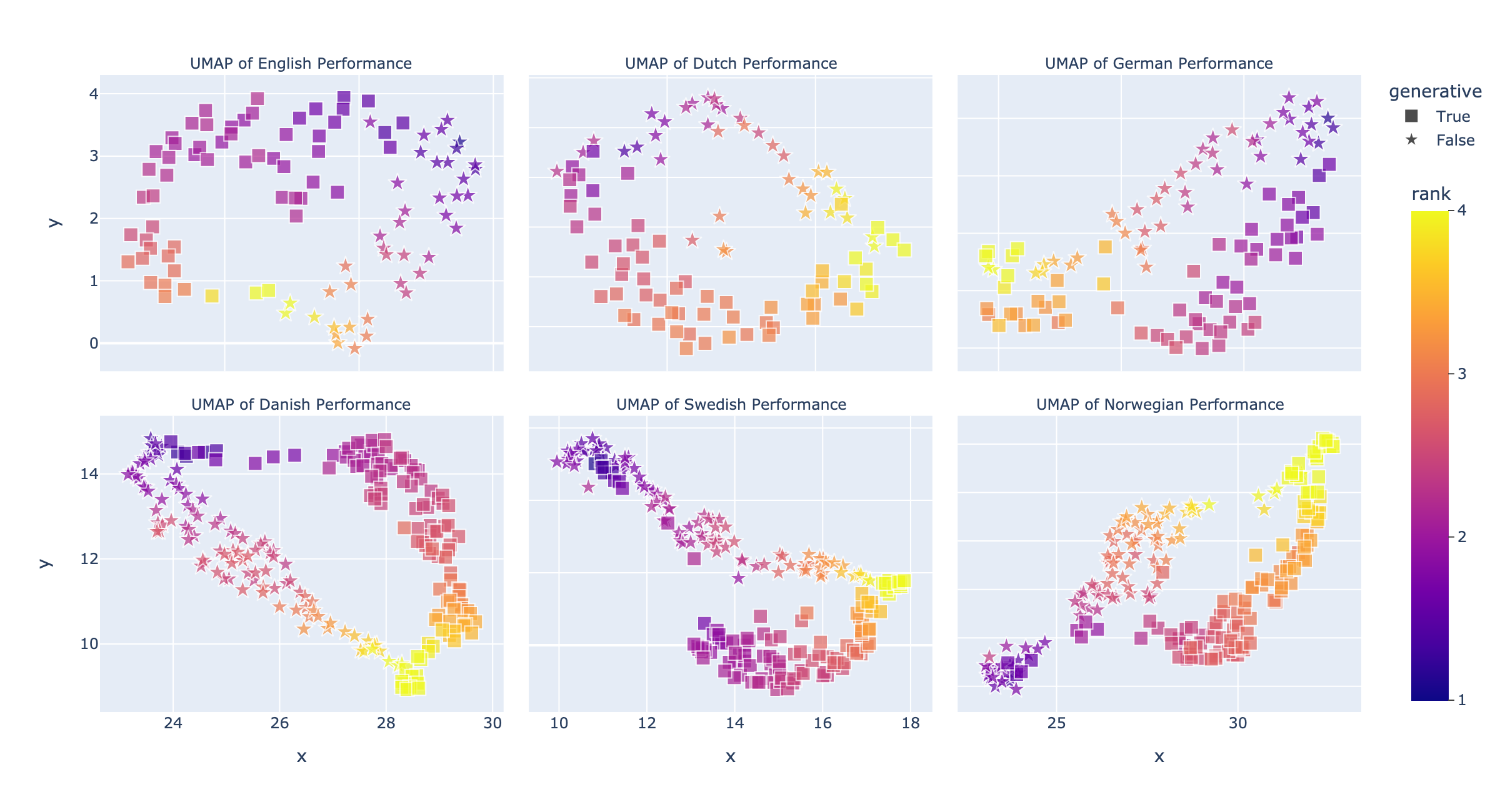}
    \caption{UMAP plots of the models on the ScandEval leaderboards.}
    \label{fig:umap}
\end{figure*}

Firstly, we analyse whether the score distribution across the four NLU tasks is
different for the encoder and decoder models. This is done by applying a UMAP
\cite{mcinnes2018umap} to the results of a given leaderboard, which is a dimensionality
reduction method that both takes into account the global and local structure of the
underlying data - it can thus be viewed as a middle ground between a principal component
analysis \cite{pearson1901liii} and a t-distributed stochastic neighbour embedding
\cite{hinton2002stochastic}. The resulting reduction thus contains a single
two-dimensional representation of each model. UMAP plots for the English, Danish,
Swedish, Norwegian, German and Dutch leaderboards can be found in Figure~\ref{fig:umap},
where we also mark the mean rank score for each model, as well as whether the model is
generative.

We see that the worst and best performing models have similar distributions,
irrespective of whether they are generative or not. However, we also note that the
rest of encoder and decoder models follow different ``paths'' in the UMAP space,
leading to our hypothesis that the different architectures have different task
preferences.

\begin{figure}[h]
    \centering
    \includegraphics[width=.4\textwidth]{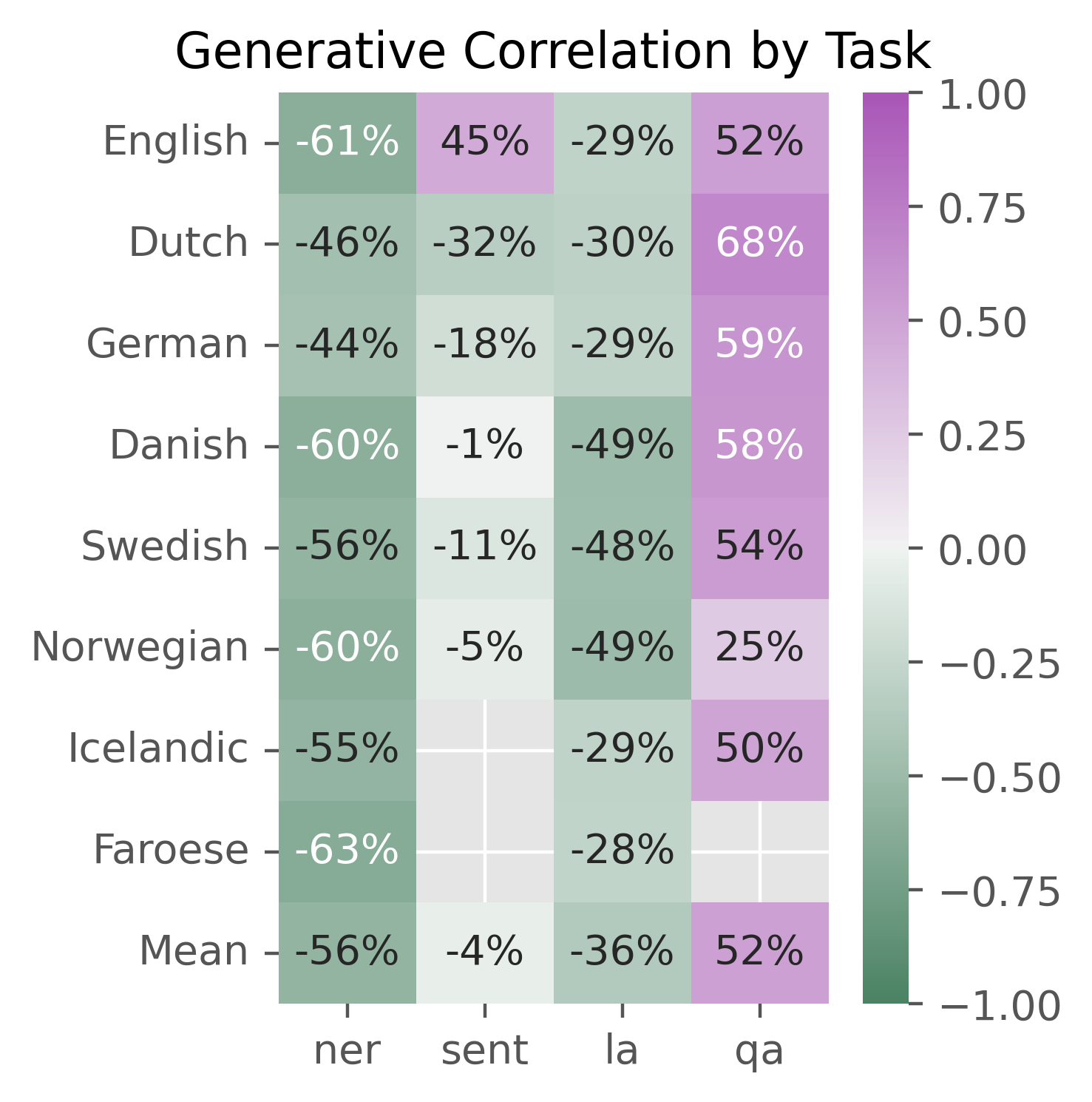}
    \caption{The correlation between a model being generative and its performance on
    the four NLU tasks.}
    \label{fig:generative-correlation}
\end{figure}

In Figure~\ref{fig:generative-correlation} we show the correlation between a model
being generative and its performance on the four NLU tasks. We see that being
generative is a strong predictor for good question answering performance, as well as
poor named entity recognition and linguistic acceptability performance. The correlation
is weaker for sentiment classification and varies across languages. We also see that
these findings seem to generalise across languages, both high- and low-resource. The
large question answering performance persists for non-instruction-tuned decoder models
(see the leaderboards at \leaderboardurl), showing a likely side-effect of the
pre-training algorithm or the architecture of decoder models making them better at this
task. We also note that generative models perform substantially better at the English
sentiment classification dataset SST5 compared to the other sentiment classification
datasets - we will return to this in the discussion.

\section{Discussion}
Having a good mean rank score is not the only thing that matters when choosing a model
for a given task. Model size, inference speed and whether the model has publicly
available weights are all important factors to consider. For this reason, we also
include these metadata in the leaderboard, and we encourage the community to consider
these factors when choosing a model for a given task.

Some of the datasets in the benchmark are translations of American datasets, which we
acknowledge is not ideal and encourage the development of gold-standard replacements of
these. This concerns the Dutch question answering dataset, which is machine translated,
as well as the Danish and Swedish question answering datasets, where the questions and
answers have been manually translated. Manual translations are typically better than
machine translations, but it nevertheless means that the content is biased towards
questions pertinent to the American context. Some datasets are furthermore missing. This
concerns Icelandic and Faroese sentiment analysis, as well as Faroese question
answering. Efforts are currently underway to remedy this.

Lastly, we note that the English sentiment classification dataset SST5 is the only
dataset where generative models perform substantially better than encoder models. We
speculate that this is either due to the dataset simply being significantly easier than
the others, or that the test data has leaked into the pretraining datasets of the
generative models. The dataset is part of the FLAN collection \cite{weifinetuned}, which
is for instance included in the Dolma dataset \cite{soldaini2024dolma}, which is used to
pretrain the OLMo model \cite{groeneveld2024olmo}, being one of the generative models
that is performing very well on this dataset. Leakage is therefore possible, and we
encourage the community to investigate this further.

\section{Conclusion}
We have extended the ScandEval benchmark to include the evaluation of decoder models, as
well as including three new languages: German, Dutch and English. From the analysis of
the corresponding results we found that encoder models can achieve significantly better
NLU performance than decoder models despite having orders of magnitude fewer parameters,
but that this varies between languages. We have also shown that being generative is
strongly correlated with both good question answering performance and poor performance
for named entity recognition and linguistic acceptability. Our analysis showed that the
``path'' from the worst to the best-performing models in the UMAP space is different for
encoder and decoder models, indicating an architecture-specific task-preference.

\section*{Ethics Statement}
We have made efforts towards making the evaluation as fair and unbiased as possible,
both through our selection of the datasets in the benchmark as well as through our
choice of aggregation method of the scores. However, we have not conducted extensive
bias analyses on the individual evaluation datasets.

\section*{Acknowledgements}
This work has received funding by the European Union’s Horizon 2023 Research and
Innovation Actions, as part of the Artificial Intelligence and Robotics programme, for
the project ``TrustLLM'' (grant agreement number 101135671). Furthermore, this work
reflects only the authors’ view and the European Research Executive Agency (REA) is not
responsible for any use that may be made of the information it contains.

\newpage
\bibliographystyle{acl_natbib}
\bibliography{nodalida2025}

\appendix
\onecolumn

\section{Prompts}
\label{apx:prompts}

\begin{table}[H]
    \centering
    \footnotesize
    \begin{tblr}{Q[3cm,valign=m]|Q[7cm,valign=m]X[j,valign=m]}
        \hline
        \textbf{Dataset} & \textbf{Prefix Prompt} & \textbf{Example Prompt} \\
        \hline
        DANSK & Følgende er sætninger og JSON-ordbøger med de navngivne enheder, som forekommer i den givne sætning. & Sætning: [text]\newline Navngivne enheder: [label] \\
        SUC 3.0 & Följande är meningar och JSON-ordböcker med de namngivna enheter som förekommer i den givna meningen. & Mening: [text]\newline Namngivna entiteter: [label] \\
        NorNE-nb\newline NorNE-nn & Følgende er fraser og JSON-ordbøker med de navngitte enhetene som forekommer i den gitte frasen. & Frase: [text]\newline Navngitte enheter: [label] \\
        MIM-GOLD-NER & Eftirfarandi eru setningar ásamt JSON lyklum með nefndum einingum sem koma fyrir í setningunum. & Setning: [text]\newline Nefndar einingar: [label] \\
        FoNE & Her eru nakrir setningar og nakrar JSON orðabøkur við nevndar eindir, sum eru í setningunum. & Setningur: [text]\newline Nevndar eindir: [label] \\
        GermEval & Es folgen Sätze und JSON-Wörterbücher mit den benannten Entitäten, die in der angegebenen Phrase vorkommen. & Satz: [text]\newline Benannte Entitäten: [label] \\
        CoNLL-nl & Hieronder staan zinnen en JSON woordenboeken met de genoemde entiteiten die voorkomen in de gegeven zin. & Zin: [text]\newline Genoemde entiteiten: [label] \\
        CoNLL-en & Below are sentences and JSON dictionaries with the named entities that occur in the given sentence. & Sentence: [text]\newline Named entities: [label] \\
        \hline
    \end{tblr}
    \caption{The prompt templates used for the NER datasets.}
\end{table}

\begin{table}[H]
    \centering
    \small
    \begin{tblr}{l|Q[7cm,valign=m]X[j,valign=m]}
        \hline
        \textbf{Dataset} & \textbf{Prefix Prompt} & \textbf{Example Prompt} \\
        \hline
        Angry Tweets & Følgende er tweets og deres sentiment, som kan være 'positiv', 'neutral' eller 'negativ'. & Tweet: [text]\newline Sentiment: [label] \\
        SweReC & Följande är recensioner och deras sentiment, som kan vara 'positiv', 'neutral' eller 'negativ'. & Recension: [text]\newline Sentiment: [label] \\
        NoReC & Følgende er anmeldelser og deres sentiment, som kan være 'positiv', 'nøytral' eller 'negativ'. & Anmeldelse: [text]\newline Sentiment: [label] \\
        SB10k & Im Folgenden sind Tweets und ihre Stimmung aufgeführt, die 'positiv', 'neutral' oder 'negativ' sein kann. & Tweet: [text]\newline Stimmungslage: [label] \\
        Dutch Social & Hieronder staan tweets en hun sentiment, dat 'positief', 'neutraal' of 'negatief' kan zijn. & Tweet: [text]\newline Sentiment: [label] \\
        SST5 & The following are tweets are their sentiment, which can be 'positive', 'neutral' or 'negative'. & Tweet: [text]\newline Sentiment: [label] \\
        \hline
    \end{tblr}
    \caption{The prompt templates used for the sentiment classification datasets.}
\end{table}

\begin{table}[H]
    \centering
    \small
    \begin{tblr}{Q[2cm,valign=m]|Q[7cm,valign=m]X[j,valign=m]}
        \hline
        \textbf{Dataset} & \textbf{Prefix Prompt} & \textbf{Example Prompt} \\
        \hline
        ScaLA-da & Følgende er sætninger og om de er grammatisk korrekte. & Sætning: [text]\newline Grammatisk korrekt: [label] \\
        ScaLA-sv & Följande är meningar och huruvida de är grammatiskt korrekta. & Mening: [text]\newline Grammatisk korrekt: [label] \\
        ScaLA-nb\newline ScaLA-nn & Følgende er setninger og hvorvidt de er grammatisk korrekte. & Setning: [text]\newline Grammatisk korrekt: [label] \\
        ScaLA-is & Eftirfarandi eru setningar og hvort þær eru málfræðilega réttar. & Setning: [text]\newline Málfræðilega rétt: [label] \\
        ScaLA-fo & Hetta eru nakrir setningar og um teir eru mállæruliga rættir. & Setningur: [text]\newline Mállæruliga rættur: [label] \\
        ScaLA-de & Die folgenden Sätze und ob sie grammatikalisch korrekt sind. & Satz: [text]\newline Grammatikalisch richtig: [label] \\
        ScaLA-nl & Hieronder staan zinnen en of ze grammaticaal correct zijn. & Zin: [text]\newline Grammaticaal correct: [label] \\
        ScaLA-en & The following are sentences and whether they are grammatically correct. & Sentence: [text]\newline Grammatically correct: [label] \\
        \hline
    \end{tblr}
    \caption{The prompt templates used for the linguistic acceptability datasets.}
\end{table}

\begin{table}[H]
    \centering
    \small
    \begin{tblr}{l|Q[5cm,valign=m]X[j,valign=m]}
        \hline
        \textbf{Dataset} & \textbf{Prefix Prompt} & \textbf{Example Prompt} \\
        \hline
        ScandiQA-da & Følgende er tekster med tilhørende spørgsmål og svar. & Tekst: [text]\newline Spørgsmål: [question]\newline Svar med maks. 3 ord: [label] \\
        ScandiQA-sv & Her følger tekster med tilhørende spørsmål og svar. & Tekst: [text]\newline Spørsmål: [question]\newline Svar på maks 3 ord: [label] \\
        NorQuAD & Nedan följer texter med tillhörande frågor och svar. & Text: [text]\newline Fråga: [question]\newline Svar på max 3 ord: [label] \\
        NQiI & Eftirfarandi eru textar með tilheyrandi spurningum og svörum. & Texti: [text]\newline Spurning: [question]\newline Svaraðu með að hámarki 3 orðum: [label] \\
        GermanQuAD & Im Folgenden finden Sie Texte mit den dazugehörigen Fragen und Antworten. & Text: [text]\newline Fragen: [question]\newline Fragen Antwort in maximal 3 Wörtern: [label] \\
        SQuAD-nl & Hieronder volgen teksten met bijbehorende vragen en antwoorden. & Tekst: [text]\newline Vraag: [question]\newline Antwoord in max 3 woorden: [label] \\
        SQuAD & The following are texts with accompanying questions and answers. & Text: [text]\newline Question: [question]\newline Answer in max 3 words: [label] \\
        \hline
    \end{tblr}
    \caption{The prompt templates used for the question answering datasets.}
\end{table}

\begin{table}[H]
    \centering
    \small
    \begin{tblr}{l|cccccccc}
        \hline
        \textbf{Language} & \textbf{PER} & \textbf{LOC} & \textbf{ORG} & \textbf{Negative} & \textbf{Neutral} & \textbf{Positive} & \textbf{Correct} & \textbf{Incorrect} \\
        \hline
        Danish & Person & Sted & Organisation & Negativ & Neutral & Positiv & Ja & Nej \\
        Swedish & Person & Plats & Organisation & Negativ & Neutral & Positiv & Ja & Nej \\
        Norwegian & Person & Sted & Organisasjon & Negativ & Nøytral & Positiv & Ja & Nei \\
        Icelandic & Einstaklingur & Staðsetning & Stofnun & & & & Já & Nei \\
        Faroese & Persónur & Staður & Felagsskapur & & & & Ja & Nei \\
        German & Person & Ort & Organisation & Positiv & Neutral & Negativ & Ja & Nein \\
        Dutch & Persoon & Locatie & Organisatie & Positief & Neutraal & Negatief & Ja & Nee \\
        English & Person & Location & Organization & Negative & Neutral & Positive & Yes & No \\
        \hline
    \end{tblr}
    \caption{The label conversions for the datasets.}
\end{table}

\section{Mean Rank Score Algorithm}
\label{apx:mean_rank_score}

\begin{algorithm}[H]
\caption{The mean rank score algorithm.}
\label{alg:mean_rank_score}
\begin{algorithmic}[1]
    \State \textbf{Input:} $M$ models, $D$ datasets, $S_{m,d,i}$ the $i$'th raw score of model $m$ on dataset $d$
    \State \textbf{Output:} The mean rank score of each model
    \State $R\gets\textrm{matrix of size }M\times D$
    \For {$\hat d \in D$}
        \State $\textrm{sortedModels}\gets\textrm{argsort}_m\sfrac{1}{10}\cdot\Sigma_i S_{m,\hat d,i}$
        \For {$\hat m\in\textrm{sortedModels}$}
            \If {$\hat m$ is the first model}
                \State $\rho\gets 1$
                \State $\textrm{previousScores}\gets S_{\hat m,\hat d}$
            \Else
                \State $\textrm{pValue}\gets\textrm{one-tailed Welch's t-test}(\textrm{previousScores}, S_{\hat m,\hat d})$
                \If {$\textrm{pValue}<0.05$}
                    \State $\Delta\gets\textrm{mean}(S_{\hat m,\hat d})-\textrm{mean}(\textrm{previousScores})$
                    \State $\rho\gets\rho+\sfrac{\Delta}{\textrm{standardDeviation}_m S_{m,\hat d}}$
                    \State $\textrm{previousScores}\gets S_{\hat m,\hat d}$
                \EndIf
            \EndIf
            \State $R_{\hat m, \hat d}\gets\rho$
        \EndFor
    \EndFor
    \State $\textrm{output}\gets\sfrac{1}{D}\cdot\Sigma_d R_{m,d}$
\end{algorithmic}
\end{algorithm}

\section{Leaderboard Excerpts}
\label{apx:leaderboard_excerpts}

\begin{table}[H]
    \centering
    \begin{tabular}{l|cc}
        \hline
        \textbf{Model ID} & \textbf{Decoder} & \textbf{Score ($\downarrow$)} \\
        \hline
        AI-Sweden-Models/Llama-3-8B-instruct & \tick & 1.07 \\
        gpt-4-0613 & \tick & 1.28 \\
        four-two-labs/lynx-micro & \tick & 1.29 \\
        AI-Sweden-Models/roberta-large-1350k & \cross & 1.33 \\
        meta-llama/Meta-Llama-3-70B & \tick & 1.34 \\
        gpt-4o-2024-05-13 & \tick & 1.35 \\
        gpt-4-1106-preview & \tick & 1.37 \\
        KBLab/megatron-bert-large-swedish-cased-165k & \cross & 1.42 \\
        AI-Sweden-Models/bert-large-nordic-pile-1M-steps & \cross & 1.48 \\
        KB/bert-base-swedish-cased & \cross & 1.48 \\
        intfloat/multilingual-e5-large & \cross & 1.49 \\
        ltg/norbert3-large & \cross & 1.49 \\
        gpt-3.5-turbo-0613 & \tick & 1.57 \\
        upstage/SOLAR-10.7B-v1.0 & \cross & 1.96 \\
        timpal0l/Mistral-7B-v0.1-flashback-v2 & \tick & 2.02 \\
        mistralai/Mistral-7B-v0.1 & \tick & 2.17 \\
        \hline
    \end{tabular}
    \caption{Excerpt of the Swedish ScandEval leaderboard.}
    \label{tab:swedish-leaderboard}
\end{table}

\begin{table}[H]
    \centering
    \begin{tabular}{l|cc}
        \hline
        \textbf{Model ID} & \textbf{Decoder} & \textbf{Score ($\downarrow$)} \\
        \hline
        ltg/norbert3-large & \cross & 1.12 \\
        NbAiLab/nb-bert-large & \cross & 1.19 \\
        ltg/norbert3-base & \cross & 1.32 \\
        AI-Sweden-Models/roberta-large-1350k & \cross & 1.38 \\
        gpt-4-0613 & \tick & 1.49 \\
        danish-foundation-models/encoder-large-v1 & \cross & 1.54 \\
        google/rembert & \cross & 1.57 \\
        meta-llama/Meta-Llama-3-70B & \tick & 1.59 \\
        gpt-4-1106-preview & \tick & 1.62 \\
        gpt-4o-2024-05-13 & \tick & 1.67 \\
        microsoft/mdeberta-v3-base & \cross & 1.67 \\
        gpt-3.5-turbo-0613 & \tick & 2.03 \\
        upstage/SOLAR-10.7B-v1.0 & \cross & 2.38 \\
        meta-llama/Meta-Llama-3-8B & \tick & 2.45 \\
        mistralai/Mistral-7B-v0.1 & \tick & 2.79 \\
        AI-Sweden-Models/Llama-3-8B-instruct & \tick & 2.82 \\
        \hline
    \end{tabular}
    \caption{Excerpt of the Norwegian ScandEval leaderboard.}
    \label{tab:norwegian-leaderboard}
\end{table}

\begin{table}[H]
    \centering
    \begin{tabular}{l|cc}
        \hline
        \textbf{Model ID} & \textbf{Decoder} & \textbf{Score ($\downarrow$)} \\
        \hline
        vesteinn/FoBERT	& \cross & 1.00 \\
        microsoft/mdeberta-v3-base & \cross & 2.01 \\
        vesteinn/ScandiBERT-no-faroese & \cross & 2.66 \\
        gpt-4-1106-preview & \tick & 2.69 \\
        google/rembert & \cross & 3.35 \\
        ltg/norbert3-base & \cross & 3.36 \\
        gpt-4-0613 & \tick & 3.46 \\
        mideind/IceBERT-large & \cross & 3.96 \\
        meta-llama/Meta-Llama-3-70B & \tick & 4.16 \\
        gpt-3.5-turbo-0613 & \tick & 4.23 \\
        gpt-4o-2024-05-13 & \tick & 4.69 \\
        mistralai/Mistral-7B-v0.1 & \tick & 4.84 \\
        \hline
    \end{tabular}
    \caption{Excerpt of the Faroese ScandEval leaderboard.}
    \label{tab:faroese-leaderboard}
\end{table}

\begin{table}[H]
    \centering
    \begin{tabular}{l|cc}
        \hline
        \textbf{Model ID} & \textbf{Decoder} & \textbf{Score ($\downarrow$)} \\
        \hline
        deepset/gbert-large & \cross & 1.17 \\
        FacebookAI/xlm-roberta-large & \cross & 1.34 \\
        google/rembert & \cross & 1.34 \\
        gpt-4-0613 & \tick & 1.49 \\
        intfloat/multilingual-e5-large & \cross & 1.49 \\
        meta-llama/Meta-Llama-3-70B & \tick & 1.57 \\
        gwlms/deberta-base-dewiki-v1 & \cross & 1.61 \\
        gpt-4-1106-preview & \tick & 1.65 \\
        upstage/SOLAR-10.7B-v1.0 & \cross & 1.67 \\
        microsoft/mdeberta-v3-base & \cross & 1.74 \\
        gpt-4o-2024-05-13 & \tick & 1.82 \\
        deepset/gbert-base & \cross & 1.83 \\
        gpt-3.5-turbo-0613 & \tick & 1.95 \\
        occiglot/occiglot-7b-de-en-instruct & \tick & 2.02 \\
        meta-llama/Meta-Llama-3-8B & \tick & 2.02 \\
        mistralai/Mistral-7B-v0.1 & \tick & 2.15 \\
        \hline
    \end{tabular}
    \caption{Excerpt of the German ScandEval leaderboard.}
    \label{tab:german-leaderboard}
\end{table}

\begin{table}[H]
    \centering
    \begin{tabular}{l|cc}
        \hline
        \textbf{Model ID} & \textbf{Decoder} & \textbf{Score ($\downarrow$)} \\
        \hline
        intfloat/multilingual-e5-large & \cross & 1.42 \\
        gpt-4-0613 & \tick & 1.69 \\
        DTAI-KULeuven/robbert-2022-dutch-base & \cross & 1.78 \\
        pdelobelle/robbert-v2-dutch-base & \cross & 1.81 \\
        meta-llama/Meta-Llama-3-70B & \tick & 1.84 \\
        gpt-4-1106-preview & \tick & 1.98 \\
        intfloat/multilingual-e5-base & \cross & 2.00 \\
        FacebookAI/xlm-roberta-large & \cross & 2.10 \\
        gpt-4o-2024-05-13 & \tick & 2.14 \\
        microsoft/mdeberta-v3-base & \cross & 2.25 \\
        gpt-3.5-turbo-0613 & \tick & 2.27 \\
        google/rembert & \cross & 2.29 \\
        upstage/SOLAR-10.7B-v1.0 & \cross & 2.33 \\
        meta-llama/Meta-Llama-3-8B & \tick & 2.54 \\
        mistralai/Mistral-7B-v0.1 & \tick & 2.77 \\
        \hline
    \end{tabular}
    \caption{Excerpt of the Dutch ScandEval leaderboard.}
    \label{tab:dutch-leaderboard}
\end{table}

\end{document}